\documentclass[a4paper,twoside]{article}

\usepackage{epsfig}
\usepackage{subcaption}
\usepackage{calc}
\usepackage{amssymb}
\usepackage{amstext}
\usepackage{amsmath}
\usepackage{amsthm}
\usepackage{multicol}
\usepackage{pslatex}
\usepackage{apalike}
\usepackage{algorithm2e}
\usepackage[bottom]{footmisc}
\usepackage{xcolor}
\usepackage{SCITEPRESS}

\newcommand{\ie}{i.e. }

\newcommand{\todo}[1]{}
\renewcommand{\todo}[1]{\textcolor{red}{@TODO: {#1}}}

\begin{document}

\title{Learned Fusion: 3D Object Detection using\\Calibration-Free Transformer Feature Fusion}

\author{\authorname{Michael Fürst\sup{1,2}\orcidAuthor{0000-0001-6647-5031}, Rahul Jakkamsetty\sup{2}\orcidAuthor{0009-0000-0711-229X}, René Schuster\sup{1,2}\orcidAuthor{0000-0001-7055-9254} and Didier Stricker\sup{1,2}\orcidAuthor{0000-0002-5708-6023}}
\affiliation{\sup{1}Augmented Vision, RPTU - University of Kaiserslautern-Landau, Kaiserslautern, Germany}
\affiliation{\sup{2}DFKI - German Research Center for Artificial Intelligence, Kaiserslautern, Germany}
\email{\{firstname.lastname\}@dfki.de}
}


\abstract{
  The state of the art in 3D object detection using sensor fusion heavily relies on calibration quality, which is difficult to maintain in large scale deployment outside a lab environment.
  We present the first calibration-free approach for 3D object detection.
  Thus, eliminating the need for complex and costly calibration procedures.
  Our approach uses transformers to map the features between multiple views of different sensors at multiple abstraction levels.
  In an extensive evaluation for object detection, we not only show that our approach outperforms single modal setups by 14.1\% in BEV mAP, but also that the transformer indeed learns mapping.
  By showing calibration is not necessary for sensor fusion, we hope to motivate other researchers following the direction of calibration-free fusion.
  Additionally, resulting approaches have a substantial resilience against rotation and translation changes.
}

\onecolumn \maketitle \normalsize \setcounter{footnote}{0} \vfill

\section{\uppercase{Introduction}}
\label{sec:introduction}

Environment perception is one of the pillars of advances in automated driving.
Specifically, 3D object detection is critical, as knowing the position of objects in the world relative to the ego vehicle is needed for path planning and avoiding collisions.

As mission critical goals of automated driving are safety and redundancy, this must be reflected in all approaches.
Thus, many object detectors apply sensor fusion to increase the average precision over single modal approaches.
For example the lidar samples a far away object only with as few as 1-7 points, rendering it hardly detectable from lidar.
In the camera image, the same object typically spans an area of more than $30 \times 30$ pixels and therefore can be recognized.

The current state of the art in 3D object detection uses calibration information in the form of a transform- and projection-matrix.
Since features are projected from one view to the other, the calibration needs to be very precise, as errors in the calibration directly impact the quality of the predictions of the model.
For example an angular error of 1° results in a misalignment of 0.7m at a distance of 40 meters, which is equal to the size of the typical bounding box of a pedestrian.

Whilst in benchmark datasets high quality calibration is given, it is difficult to obtain and maintain high quality calibration at a production scale.
Since there is variance in production, calibration must be done per car.
Furthermore, during the lifetime of a vehicle deformations and thus changes in the calibration can happen due to heat, vibration and even replacement of sensors or defective parts.
For calibration, typically special environments with markers are required.
If a vehicle has to be regularly re-calibrated in a special environment, this poses a substantial challenge to automated driving at scale.

\begin{figure}
  \includegraphics[width=0.48\textwidth]{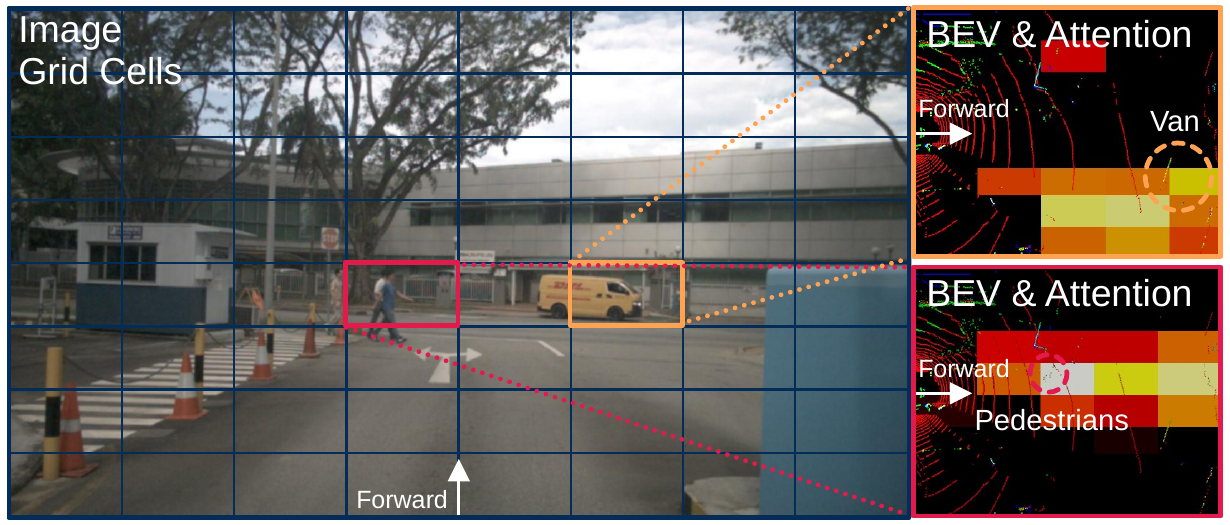}
  \caption{The attention for the highlighted grid cells (left) are overlayed over the BEV lidar (right). Brighter values (yellow) mean more attention and dark (black) is no attention. The cone corresponding to the grid cell has a high attention, while the rest has low attention. Our calibration-free approach learns the correspondence of image and bird's-eye-view (BEV).}
  \label{fig:attention_vis}
\end{figure}

Instead of improving calibration or introducing continuous calibration during operation of the vehicle adding complexity, we see the solution in eliminating the need for calibration.
Thus, we propose and contribute:
\begin{itemize}
  \item The new category of approaches doing calibration-free sensor fusion for object detection,
  \item a concrete implementation using transformers, exploiting the characteristics of self-attention,
  \item an analysis of the effectiveness, showing that the fusion can actually be learned (see Figure~\ref{fig:attention_vis}).
\end{itemize}

\section{\uppercase{Related Work}}

Object detection for autonomous vehicles is a very actively researched field with many approaches.
Thus, we focus on the approaches, which we consider the most influential and closely related to our work.
We further categorize the work in two groups: 3D object detectors and transformers in detection.

\subsection{3D Object Detectors}

\textbf{Monocular} (RGB-only) 3D detection is popular due to the cheap price of cameras. Approaches like MergeBox~\cite{gahlert2018mb}, \cite{mousavian20173D} and Direct 3D Detection~\cite{weber2019direct} are some of the first to push the performance envelope for monocular detection.
These early approaches have difficulties with the depth perception due to the inherent depth ambiguity of monocular data.

The current state of the art research like \cite{zong2023temporal}, \cite{wang2023exploring} and \cite{liu2023sparsebev} focuses on temporal information and its efficient use for detection to partially overcome depth ambiguity.
However, their prediction quality still lacks behind approaches using lidar.

PseudoLiDAR~\cite{wang2019pseudo} discovered, that convolution of depth information in the camera view is sub-optimal and introduced a BEV lidar-like representation for the depth map called PseudoLiDAR.
Leading to our choice of BEV features as the primary source for BEV object detection.

\textbf{LiDAR-based} detection overcomes the depth ambiguity and achieves much higher mean average precision (mAP) than monocular approaches.
PointPillars~\cite{lang2019pointpillars} is one of the first approaches for automated driving to combine point cloud processing with efficient CNN backbones.

CenterPoint~\cite{yin2021center} successfully extends the concept of CenterNet~\cite{duan2019centernet} to 3D object detection.
At the core, the bounding box prediction is separated into two tasks: Predicting the center point of the box and predicting the size and orientation.
The prediction of the center point is done via a heat map per object class and allows for easy and precise localization of objects independent of their orientation.

The current state of the art researches temporal aspects~\cite{koh2023mgtanet}, scene synthesis~\cite{zhan2023real}, focusing on hard samples~\cite{chen2023focalformer3d} and innovative kernels~\cite{chen2023focalformer3d}.
Techniques used here might be applicable in learned fusion as well in future research.

\textbf{Sensor Fusion} for detecting objects uses both lidar and RGB.
The goal of using multiple sensors is better performance and robustness of the approaches by leveraging the strengths of both sensors.
Image data is strong at initial recognition of objects, while lidar is strong at precise localization of objects in 3D.

F-PointNet~\cite{qi2018frustum} first uses a 2D detector to then crop a frustum in the point cloud using the calibration and predict the 3D object in that frustums point cloud.
PointPainting~\cite{vora2020pointpainting} predicts segmentation masks using the camera and then colorizes the point cloud to predict the boxes there.

AVOD~\cite{ku2018joint} projects anchors into the views using the calibration to allow them to crop the features and concatenate them.
Then, proposals are predicted which are projected into the views again.
Finally, the cropped fused features are used for precise bounding box prediction.

Differently, LRPD~\cite{furst2020lrpd} first uses instance segmentation to generate proposals and then projects the proposals to the views and fuses similar to AVOD.
It leverages RGB for initial recognition and then both sensors for the fine localization.

Following the idea of explicit handling of sensors, BEV Fusion~\cite{liu2023bevfusion} transforms the camera features efficiently to the bird's-eye-view and then stacks them with the lidar features to predict the bounding boxes.
The mapping is pre-computed using the calibration.

All current sensor fusion approaches in 3D object detection use calibration matrices.
The most common use is to map information from one view to another.

\subsection{Transformers in Detection}

Transformers as introduced by~\cite{vaswani2017attention} use self-attention at the core.
The self-attention is a dot product of the query and key vectors to create a weight matrix called attention matrix.
The attention matrix is then multiplied with the values resulting in a weighted sum of the values.

Detection Transformer (DeTr)~\cite{carion2020end} very successfully applied transformers to object detection by using the features of the convolutional encoder as inputs to a transformer encoder and decoder architecture.
In contrast to common practice in detection, DeTr predicts a set of bounding boxes instead of predicting boxes for each pixel or grid cell in the image.
This makes the approach very general, but more difficult to train.

TransFusion~\cite{bai2022transfusion} applies the ideas of DeTr to sensor fusion.
It encodes the features using regular backbones and then uses queries to decode the bounding boxes.
The queries are generated from a heat map for the lidar initially and then in a second step by projecting the center of a query to 2D in the spatially modulated cross attention introduced by the approach.

Cross-Modal-Transformer (CMT)~\cite{yan2023cross} is one of the best published approaches on nuScenes~\cite{caesar2020nuscenes}.
The core concept is very similar to DeTr.
It first extracts the features from camera and lidar using encoders.
Then it concatenates the features and applies a transformer decoder.
The queries are computed from 3D points which are projected to the respective views of the sensors.
Thus enabling the correlation with the position embedding of the respective views.
We found in preliminary experiments, that the model does not converge if the calibration is omitted from this step.

Finally, TransFuser~\cite{chitta2023transfuser} uses transformers to fuse the features from the different views for learning navigation in a simulated environment.
The approach does not use calibration in its fusion, but it is limited to predicting a global output: The next N waypoints.
The TransFusion module introduced in their approach is very flexible.
However, their overall architecture produces global features unsuited for object detection.

Even when using transformers object detectors rely on calibration to map information from one view to another.
TransFuser not needing calibration cannot be applied to object detection without substantial modification.

\section{\uppercase{Approach}}

Current state of the art approaches for sensor fusion rely on calibration.
Even approaches using transformers still use calibration for fusion.
It is very difficult to successfully train a completely calibration-free fusion approach, since "object queries might attend to visual regions unrelated to the bounding box to be predicted, leading to a long training time for the network"~\cite{bai2022transfusion}.
Indeed, naively removing calibration from existing transformer based approaches does not work in our preliminary experiments.
Hence, we focus on a simple model and approach that can manage a stable training in a timely manner to show the possibility of training such calibration-free models.

Since training transformers to map between the different views is already a hard problem, the rest of the model is kept small and straightforward.
We focus on using transformers to correlate features between the views and eliminating the calibration.

\begin{figure*}
  \includegraphics[width=\textwidth]{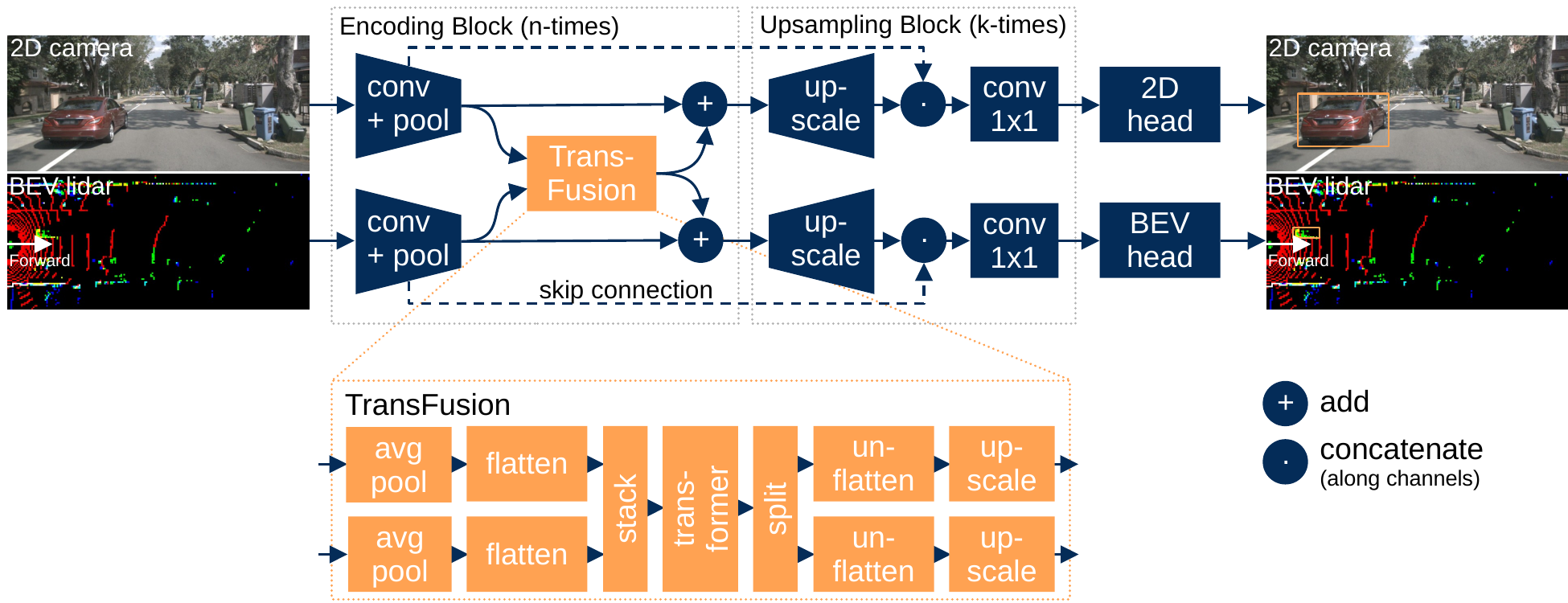}
  \caption{Our model is composed of three main components: First, the encoding block using TransFusion to fuse the features without calibration. The block is repeated $n=4$ times and the fusion (orange) is optional in each block. Second, the upsampling block increasing the resolution of the feature maps. Upsampling is also repeated $k=3$ times each time increasing the resolution by a factor of two. Last, a task-specific head predicting the bounding boxes in BEV and 2D.}
  \label{fig:architecture}
\end{figure*}

\subsection{TransFuseDet}

Giving a simplified overview over our approach, our model consists of a fusion encoder, upsampling and task-specific heads.
We use two ResNets~\cite{he2016deep} to encode lidar and camera respectively.
Transformers fuse the features at different abstraction levels, similar to TransFuser~\cite{chitta2023transfuser}.
However, unlike their approach, we keep the features of BEV and camera view separated to keep the spatial interpretation of the feature maps intact.
This allows us to do object detection.

The features which are separate but mapped via transformer fusion are then upsampled to increase resolution and finally used in two CenterNet-Style decoders for BEV object detection on the BEV features and 2D object detection on the features in camera space.
The main goal is BEV object detection, but the 2D object detection can be viewed as an auxiliary task to improve BEV performance, as we will show in an ablation study.

Figure~\ref{fig:architecture} gives an overview over our architecture and the following sections explain more details on how the transformer fusion works.

\subsection{Feature-Extraction: Multi-View Fusion via Transformers}

To extract the features, we partially follow the methodology of TransFuser~\cite{chitta2023transfuser}.
We first apply a convolutional block with pooling from the ResNets to each of the modalities.
Then, the features extracted are fused by a transformer.
However, in order to apply a transformer, the features need to be pooled, flattened and then concatenated from the different views of the sensors.
After the transformer, the features must be split into the views, reshaped and upscaled, so they can be added to the features of the respective views.

Deviating from TransFuser, the fusion step is optional after each block in the ResNet.
In the ablation study, we experimentally derive an optimal configuration.

Additionally, we introduce upsampling to increase the resolution again, as the native output resolution of the fusion encoder is too low for precise object detection.
For upsampling we simply upscale the features, concatenate them with the higher resolution features and apply 1x1 convolutions.

\subsubsection{Property of transformers allowing calibration-free fusion}

At the core of the fusion lies a transformer.
The critical component of a transformer for fusion is the attention, as it allows to correlate features independent of their spatial location.
For example, in our bird's-eye-view (BEV) representation the ego-vehicle is on the left facing towards the right as shown by the little arrow in Figure~\ref{fig:architecture}.
Thus, if we have a car which is far away in the scene, it might appear centered in the camera view, but in the lidar bird's-eye-view (BEV) it is to the far right.
Typically, calibration data would be used to compute the corresponding positions in BEV and camera view and then the features would be gathered or projected to the other view.
This is not done in calibration-free fusion.
Here, we leverage the property of the dot product in attention to correlate similar features independent of their position.

Scaled Dot-Product Attention~\cite{vaswani2017attention} consists of a dot product between the query $Q$ and the key $K$, scaled using the dimensionality $d_k$ of the keys, and then weighting the values $V$ by the resulting (softmaxed) matrix.
\begin{equation}
  \text{Attention}(Q,K,V) = \text{softmax}\left(\frac{Q \cdot K^T}{\sqrt{d_k}}\right) \cdot V
\end{equation}
The dot product computes the similarity of two vectors.
This is exactly what we need to correlate features.
For example, our convolutional encoder finds a pedestrian in the image at position $x_I$, and in the lidar at position $x_L$.
Then, the resulting features $f_{x_I}$ and $f_{x_L}$ should have a similar semantic meaning, resulting in a high value for the dot product and thus a high attention to the feature of the other view.

\subsubsection{Pooling in Fusion Reducing Computational Complexity}

Transformers have a high computational cost in image processing due to the high resolution.
To apply a transformer on a feature grid, the features must be flattened.
Thus, a vector with shape $H \times W \times C$ must be flattened to a vector of shape $(HW) \times C$ to be used as $Q=K=V$ in self-attention.
The resulting attention matrix is of shape $(HW) \times (HW) \times C$, which means $H^2 W^2 C$ values.
So for example doubling the input resolution squares the flattened vectors size and quadratically increases the size of the attention matrix, severely limiting the resolution that is practical with respect to computational cost and learnable parameters.

For fusion, we concatenate two or more flattened feature vectors before using them as $Q$, $K$ and $V$ in the self-attention.
Despite only linearly increasing complexity combined with the quadratic scaling this further increases the computational cost.

To limit cost there are two options: Reducing the resolution of the input to the model or pooling the feature maps before using them in the fusion.
Pooling the feature maps allows to keep high resolution feature maps and only sacrifice resolution in the fusion.
For object detection with CenterNet, a higher resolution is beneficial.
Higher resolution allows to distinguish two objects near each other, since only a single object per grid cell can be detected.
Additionally, the location of the maxima in the heat map can correspond better to the true mode, since quantization errors are smaller.

\subsubsection{Upsampling}

Since resolution has a big impact on the performance of the CenterNet-style decoders, increasing output feature resolution can be beneficial to the performance.
We use upsampling and combine the upsampled features with the higher resolution and lower abstraction of earlier layers of the model.
Following common convention, features are upsampled by a factor of two and combined with the features of the previous encoding block.
The combination of the features is done via a concatenation along the channel dimension followed by a 1x1 convolution.

\subsection{Detection Heads}

The detector heads follow the methodology of CenterNet~\cite{duan2019centernet} as it is a proven approach, suitable for the feature maps created by our calibration free fusion.
The heads predict a class confidence and a bounding box for each grid cell in the feature maps.
Additionally, in the BEV case we predict a yaw angle $\Theta$.

Due to the jump in the angle from 360° to 0°, directly regressing it is not possible, as the derivative would not be well defined at this jump.
Thus, we follow the convention of encoding the yaw angle $\Theta$ as a class and an offset.
Specifically, we split the angle into 8 classes representing equally sized slices of the value range from -22.5° to 337.5°. This leads to the centers of each class being at 0, 45, 90, ... 270 and 315 degrees.
The offset value range is then from -22.5° to +22.5°.

To train the heads, we use a mean squared error (MSE) over the entire grid map for the heat map loss $L_{\text{heat}}$.
Since the heat map is heavily biased towards no objects (background), we split the loss computation into background and foreground loss.
These are then summed using a weighted sum:
\begin{equation}
  L_{heat} = w_{fg} L_{mse,fg} + w_{bg} L_{mse,bg}.
\end{equation}

For the box and yaw we only apply the losses on the grid cells associated with a bounding box.
For other grid cells, no loss is computed.
The bounding box loss $L_{\text{bbox}}$ is a smooth L1-loss and for the yaw we use cross entropy for the class loss $L_{\Theta\text{-cls}}$  and smooth L1-loss for the offset regression loss $L_{\Delta\Theta}$.

Finally, for the BEV detection task the losses are combined into a single loss $L$ using a weighted sum, where $w$ denote the weights:
\begin{equation}
  L_{bev} = w_{heat} L_{heat} + w_{bbox} L_{bbox} + w_{\Theta} (L_{\Theta\textit{-cls}} + L_{\Delta\Theta}).
\end{equation}
In the case of the 2D detection task, we use the same loss without the yaw:
\begin{equation}
  L_{2D} = w_{heat} L_{heat} + w_{bbox} L_{bbox}.
\end{equation}
The two losses are then combined for the backpropagation using a weighted sum of the BEV and 2D task loss:
\begin{equation}
  L = w_{bev} L_{bev} + w_{2D} L_{2D}.
\end{equation}

\subsection{Implementation Details}

Based on experimental studies, our model is trained with an AdamW optimizer a learning rate of 1e-4 and weight decay of 1e-4.
The learning rate is reduced using exponential decay of 0.99996 every two steps.
For the loss weights we use $w_{BEV}=0.95$, $w_{2D}=0.05$, $w_{fg}=0.9$, $w_{bg}=0.1$, $w_{heat} = 10$, $w_{bbox} = 1.0$ and $w_{\Theta}=0.2$.
The model performs best without random rotations and offsets of the lidar, but horizontal mirroring is enabled for both lidar and camera.

We do not use extensive point cloud augmentation or aggregation of points from multiple frames.
This reduces our mAP significantly, but allows us to clearly show what our approach contributes instead of the influence of augmentation strategies.

\section{\uppercase{Evaluation}}

We evaluate our approach on the nuScenes dataset~\cite{caesar2020nuscenes} since it provides the required modalities, \ie lidar and camera, while having a sufficient size.
NuScenes has 40.000 annotated key frames in 1000 driving sequences, which is sufficiently large for training a transformer-based calibration-free model.
However, we noticed during preliminary experiments that larger transformer-based calibration-free models have difficulties converging on this dataset.
Thus, we kept our model small to show the possibility of calibration-free fusion.

The frames provided by nuScenes contain 6 camera images, 1 lidar scan and 5 radars.
We only use the front and rear camera, as the side cameras have many frames without objects, which yield little benefit training the model.
Additionally, using two instead of six cameras reduces training time by a factor of three.

\subsection{Advantage over Single Sensor}

\begin{table}
  \begin{center}
  \caption{Learned fusion outperforms its RGB-only or lidar-only counterpart.
  The models are as identical as possible to eliminate all other effects except for the learned fusion.}
  \label{tab:sota}
  \begin{tabular}{|l|c|c|c|c|}
  \hline
  Method & BEV mAP & 2D mAP \\
  \hline\hline
  RGB-only & - & 47.0 \\
  LiDAR-only & 37.6 & - \\
  \hline
  Learned Fusion [ours] & \textbf{42.9} & \textbf{48.7} \\
  \hline
  \end{tabular}
  \end{center}
\end{table}

\begin{table}
  \begin{center}
  \caption{Comparison of bird's-eye-view mAP between calibration-free fusion and much larger calibration-based approaches, which cannot be applied without calibration.}
  \label{tab:detection_performance}
  \begin{tabular}{|l|c|c|c|c|}
  \hline
  Method & Calib & mAP \\
  \hline\hline
  CMT~\cite{yan2023cross} & Yes & 70.4 \\
  BEVFusion~\cite{liu2023bevfusion} & Yes & 70.2 \\
  \hline
  Learned Fusion [ours] & No & 48.8 \\
  \hline
  \end{tabular}
  \end{center}
\end{table}

Our approach introduces a new category of approaches next to camera only, lidar-only and calibration-based fusion, we introduce calibration-free fusion.
Thus, comparability to existing approaches is limited.

From our approach we can simply derive a uni modal variant by removing the fusion and the sub network for the other modality.
This allows us to show the concrete benefit from calibration-free fusion over single modality.
However, due to the nature of our approach, it is not possible to add calibration.
Thus, a direct comparison of identical approaches showing the potential advantage of calibration over calibration-free is not possible.

Our approach is significantly outperforming its two derivatives using single modality.
This shows that the fusion without calibration has a clear advantage over no fusion.
Table~\ref{tab:sota} shows that fusion is 14.1\% better than the lidar-only approach in BEV detection and 3.6\% better than then camera only approach.
The smaller gap for 2D detection makes sense as camera information is generally considered sufficient for 2D detection.

However, when comparing our approach to the current best approach on nuScenes using calibration, it is evident, that there is still a significant performance gap (see Table~\ref{tab:detection_performance}).
This gap is to be expected from a completely novel approach using no calibration.
As there is little prior work to build on, many simplifications were done in this first approach.
However, we are confident, that further research adding advanced data augmentations and training strategies as well as scaling up the model can close the gap.

Without using any calibration information our approach learns to correlate the data from different views.
We believe that the calibration-free nature of our approach and the advantage over single modal approaches make this a very promising new field in sensor fusion for object detection.

\subsection{Number of Fusion Layers}\label{sec:num_fusions}

\begin{table}
  \begin{center}
  \caption{The performance impact different fusion configurations is measured. Adding fusions after the early layers of the model does not improve performance in BEV.}
  \label{tab:fuse}
  \begin{tabular}{|l|r|c|c|c|}
  \hline
  Method & Fusion & BEV mAP & 2D mAP \\
  \hline\hline
  Learned Fusion & 4 & 37.7 & 48.1 \\
  \textit{Learned Fusion} & \textit{3,4} & \textbf{42.9} & \textbf{48.7} \\
  Learned Fusion & 2,3,4 & 40.1 & 47.6 \\
  Learned Fusion & 1,2,3,4 & 39.0 & 47.0 \\
  \hline
  \end{tabular}
  \end{center}
\end{table}

\begin{figure}
  \includegraphics[width=0.48\textwidth]{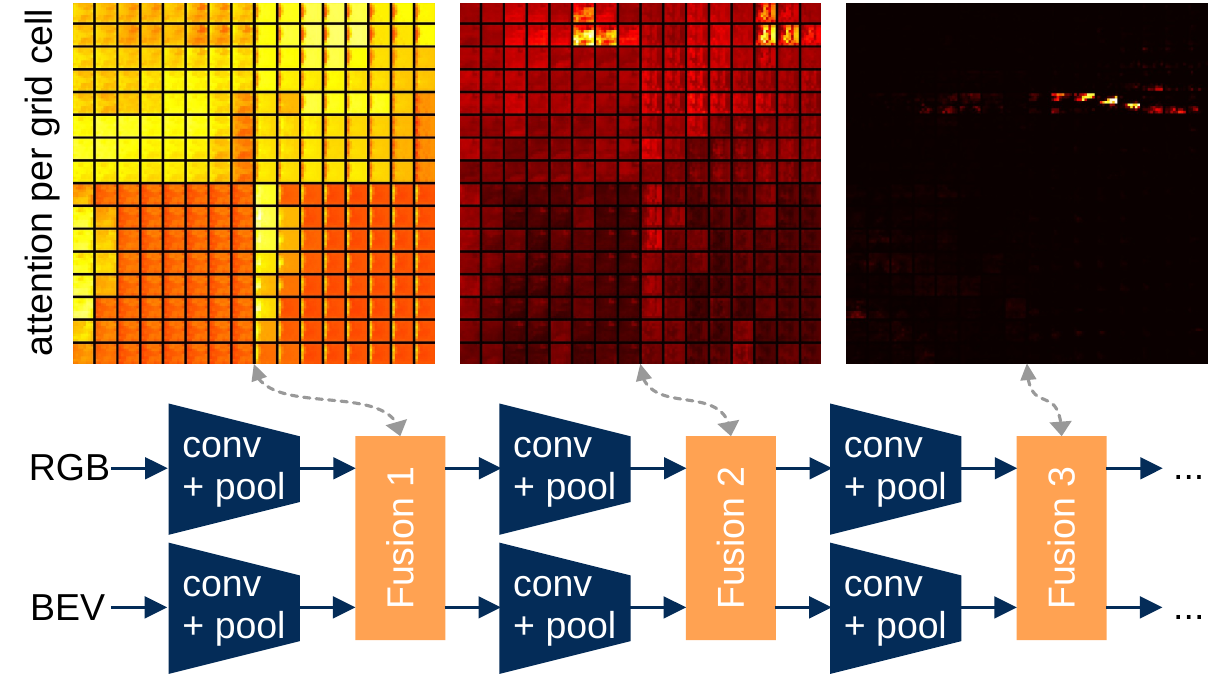}
  \caption{
    Visualizing the attention matrix, so that it corresponds to the grid cells in the image, shows that later fusions have more focus.
    Early fusions are are uniformly distributed and do not contribute  to a good fusion.}
  \label{fig:attention}
\end{figure}

Our approach allows to have a fusion in 4 optional places.
In this study we evaluate which fusions yield the best results.
We can do a fusion after each of the N last convolutional blocks in the ResNets.
For example fusions at 3 \& 4 means there is a fusion after the second last and last block in the ResNet of the two modalities. In Figure~\ref{fig:attention}, the first three fusions of a model with all fusion blocks are shown.

In Table~\ref{tab:fuse}, we can see, that more fusion layers than two does not have a positive impact on the performance of the model.
For example in the case of four fusions the BEV mAP degrades by 3.9 mAP ($\sim 9\%$) compared to two fusions.

A possible explanation is that the feature vectors correlated by our approach need a certain abstraction level.
Lower level feature maps lack the abstraction and confuse the model.
For example, a tire in an image might not be visible in the lidar at all, thus correlating these low level features does not provide much value.
Adding this fusion means the model needs to actively learn to ignore the features of a different view.
Inspecting the heat maps of the attention, we noticed that the attention for the feature maps in the earlier layers is low and unfocused, see an example in Figure~\ref{fig:attention}.

\subsection{Sensor Displacement or Rotation}

\begin{table}
  \begin{center}
  \caption{Random rotation and translation barely reduce the mAP of the model. Even extreme movement of the lidar by 5.5 meters and rotating up to 15° only has a slight impact on performance.}
  \label{tab:aug}
  \begin{tabular}{|r|r|c|c|c|}
  \hline
  Rot. & Trans. & Mirror & BEV mAP & 2D mAP \\
  \hline\hline
  0° & 0m & No & 41.1 & 47.9 \\
  \textit{0°} & \textit{0m} & \textit{Yes} & \textbf{42.9} & \textbf{48.7} \\
  0° & 0.5m & Yes & 42.3 & 46.7 \\
  15° & 0.5m & Yes & 41.3 & 47.1 \\
  15° & 5.5m & Yes & 40.5 & 45.3 \\
  \hline
  \end{tabular}
  \end{center}
\end{table}

Since our model does not use an explicit calibration, but correlates the features, it has a builtin robustness against sensor displacement and rotation.
To evaluate this, we added random rotation and translation to the input data of our model during training and testing time.

In Table~\ref{tab:aug}, it is visible, that the model with mirroring performs best.
However, adding random translation of up to 0.5m only reduces the mAP by 0.6 mAP in BEV.
Adding random rotations of up to 15° again only reduces the mAP in BEV by 1.6.
For context, calibration-based approaches expect precision in the range of 0.01m and less than 1° error.
Finally, we evaluate extreme translations up to 5.5 meters (row~5) losing 2.4 mAP over the baseline (row~2).
However, with an offset of 5.5 meters the sensor could be mounted on a vehicle next to ours, showing the extreme robustness to translation and rotation of the sensors in our approach.

These results show, that our approach has strong robustness against changes in the alignment of sensors with respect to translation and rotation.
The correlation of features via the attention is crutial for this.
We see the potential to apply this approach to multiple different vehicles without the need for modification.

\subsection{Loss and Task Weighting}

\begin{table}
  \begin{center}
  \caption{Weighting the tasks differently affects performance. BEV detection is the primary task reaching best performance at 0.95 and 0.05 weighting.}
  \label{tab:task_weight}
  \begin{tabular}{|r|r|c|c|}
  \hline
  $w_{\text{BEV}}$ & $w_{\text{2D}}$ & BEV mAP & 2D mAP \\
  \hline\hline
  0.20 & 0.80 & 38.1 & 45.3 \\
  0.50 & 0.50 & 41.8 & 49.4 \\
  0.80 & 0.20 & 42.1 & \textbf{49.7} \\
  0.90 & 0.10 & 40.7 & 47.4 \\
  \textit{0.95} & \textit{0.05} & \textbf{42.9} & \textit{48.7} \\
  0.99 & 0.01 & 41.9 & 44.0 \\
  1.00 & 0.00 & 40.5 & 0.0 \\
  \hline
  \end{tabular}
  \end{center}
\end{table}

\begin{table}
  \begin{center}
  \caption{The weighting of the different loss components has a significant impact on the performance. A high heat map weight has strong impact on the BEV mAP. Our baseline (row 3) is outperformed by a very high $w_{heat}$ (row 6).}
  \label{tab:loss_weights}
  \begin{tabular}{|r|r|c|c|c|c|}
  \hline
  $w_{\text{fg}}$ & $w_{\text{bg}}$ & $w_{\text{heat}}$ & $w_{\Theta}$ & BEV mAP & 2D mAP \\
  \hline\hline
  0.80 & 0.20 & 10 & 0.2 & 43.5 & 46.8 \\
  0.90 & 0.10 &  2 & 0.2 & 39.3 & 45.3 \\
  \textit{0.90} & \textit{0.10} & \textit{10} & \textit{0.2} & \textit{42.9} & \textit{48.7} \\
  0.90 & 0.10 & 10 & 1.0 & 44.2 & \textbf{50.7} \\
  0.90 & 0.10 & 20 & 0.2 & 42.8 & 49.4 \\
  0.90 & 0.10 & 50 & 0.2 & \textbf{48.8 }& 49.0 \\
  0.95 & 0.05 & 10 & 0.2 & 41.5 & 47.3 \\
  \hline
  \end{tabular}
  \end{center}
\end{table}

When training the approach there are hyperparemeters regarding the weighting in the loss.
We split the weighting in two categories, task weighting and loss weighting.
In task weighting we evaluate the optimal balance between the 2D and BEV detection loss with the goal of an optimal BEV mAP.
In the loss weighting we evaluate the impact of the weights for heat map foreground $w_{fg}$ and background $w_{bg}$ and the weighting of loss components using $w_{heat}$ and $w_{\Theta}$.

In Table~\ref{tab:task_weight} we can see that a higher weight to the BEV loss increases performance.
However, increasing beyond 0.95 reduces the mAP again as the 2D loss can be viewed as an auxiliary loss for the camera feature extractor.

Even without the 2D loss, the fusion model outperforms single modality.
In using only lidar achieves a 37.6 mAP while fusion without the auxiliary loss still achieves 40.5 mAP.
Fusion alone contributes substantially to the performance and the auxiliary 2D losss then further improves the effectiveness of fusion.

The evaluation of the loss weights in Table~\ref{tab:loss_weights} shows that increasing the heat map weighting in the loss has a significant impact on the mAP.
This can be explained by the fact, that the correct location of the center of a bounding box in the BEV is most important to detection, since the size variation within an object class is small.
For example, all pedestrians have almost the same size in BEV.

\subsection{Attention Map Analysis}

Besides the quantitative analysis of our approach, we visualized and analyzed the behavior of the attention maps to gain understanding.

\begin{figure}
  \includegraphics[width=0.48\textwidth]{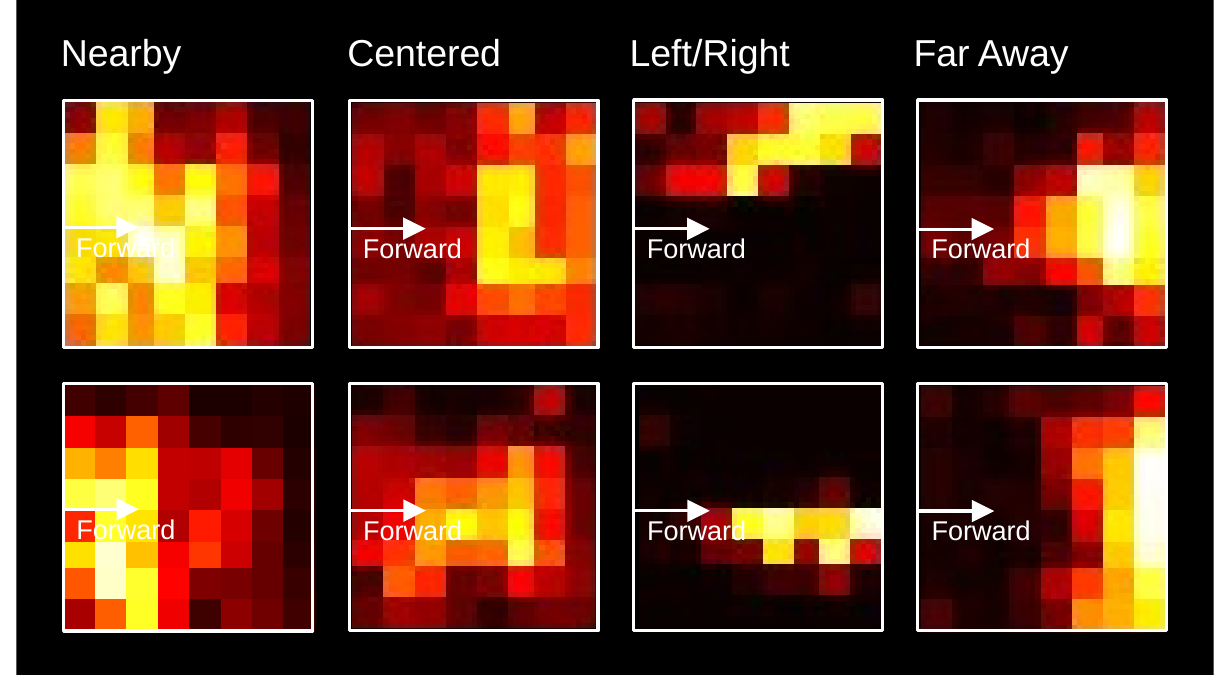}
  \caption{Attention for objects visualized in bird's-eye-view. Depending on where in the image the object is, the attention map is different. Nearby and far away objects have very different attention as well as left and right. For some objects the attention is focused, while for others it is diffuse.}
  \label{fig:attention_ablation}
\end{figure}

For the visualization we reformatted the attention map of the two flattened vectors into a grid of attention images for each cell in the feature map.
In the Figure~\ref{fig:attention} the full matrix can be seen per layer.
As described in Section~\ref{sec:num_fusions} the feature maps of earlier layers are uniform and lack focus.
Thus, for our subsequent analysis, we focused on the feature maps of fusion layers 3 and 4.

To validate that the attention has a meaningful interpretation, we analyzed the attention maps of grid cells containing objects.
In Figure~\ref{fig:attention_ablation}, we show the attention for nearby, centered, left, right and far away objects.
Generally a correlation of the attention to the region in the lidar can be seen.
However, in some cases the object attention lacks focus and is quite diffuse.
This especially happens for nearby objects covering large areas of the image.

Analyzing many of the attention maps, we noticed a few trends:
Firstly, especially the later attention layers focus their attention on cones and regions containing objects.
Secondly, the attention of regions corresponding to each other have a high attention for each other.
For example, the region at the horizon of the image and the far away region in the lidar have a high attention for each other.

We conclude from this observation, that the model learns correlating features from the regions of the different views containing the same objects.

\subsection{Qualitative results}

\begin{figure*}
  \includegraphics[width=\textwidth]{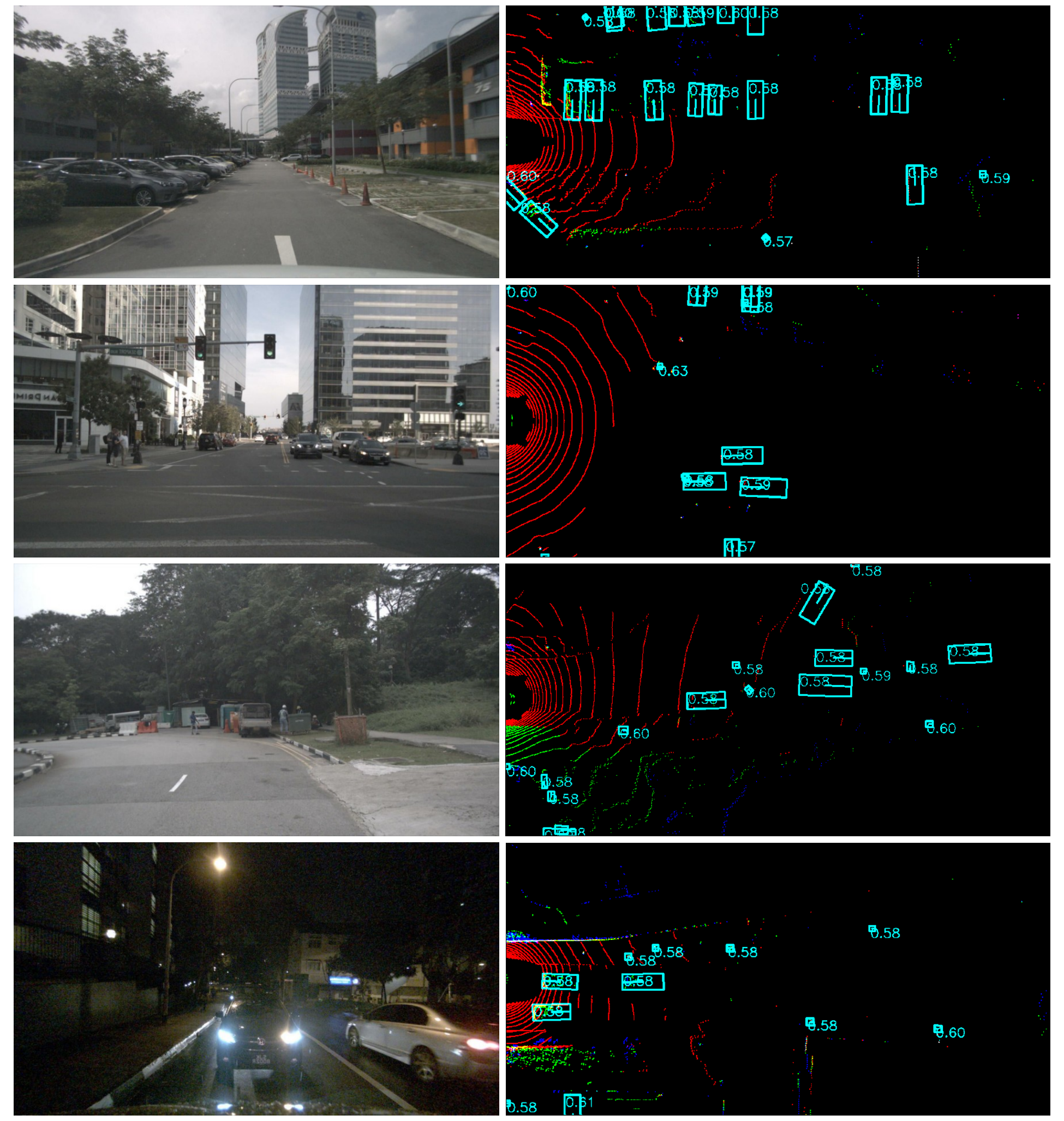}
  \caption{
    Visualization of the BEV detections predicted by our model.
    The model has good prediction for most cars, but has difficulties at the edges of the BEV lidar.
    It cannot detect all construction workers in image three, however construction workers are underrepresented in the dataset.
    At night the detection of the cars is very good despite the low visibility.
  }
  \label{fig:qualitative_results}
\end{figure*}

To validate the plausibility of our results, we visualized the BEV detections over the pointcloud input to the model.
This allows us to inspect the detections and see strengths and weaknesses of the model.

Overall, the model performance is good (see Figure~\ref{fig:qualitative_results}).
The model predicts the position and orientation of objects well.
However, in some cases the model has false positives for pedestrian objects as well as false negatives for construction workers.
The issues with construction workers are to be expected as they are an underrepresented class in the dataset.

As there is no specific shortcomings of the model apparent, our fusion seems to be quite robust.

\section{\uppercase{Conclusions}}
\label{sec:conclusion}

Overall, we introduce fully calibration-free sensor fusion using neither intrinsic nor extrinsic calibration.
Eliminating the need for complex calibration procedures in sensor fusion for object detection.
As a calibration-free sensor fusion in object detection, we present an approach using transformers for correlating features between the views and then upscaling the features to achieve the necessary resolution for precise object detection.

In the thorough evaluation, we show that calibration-free sensor fusion is a promising field.
Concretely, we show that adding calibration-free fusion increases performance over using a single modality.
Further, we found that our approach is robust against random translation and rotation, since the model correlates the features without a calibration matrix.

\subsection{Limitations and Future Research}

However, due to the complexity of the learning problem to correlate all features from the different views with each other, the model has limitations.
We identified two main limitations and propose directions for further research to eliminate them.

First, the model complexity is a problem.
The transformer for the fusion requires many of parameters and thus allows for a very limited resolution.
This leads to a gap in performance compared to the best calibration-based fusion approaches.
We expect that approaches such as deformable-attention or a query based approach like DeTr could help here.

The second limitation is the stability of the training.
During early experimentation we were experimenting with a much larger DeTr style model, but found the training to be too unstable.
We attribute this to the fact that learning a correlation between random features at the beginning of the training is highly unstable.
Thus, advanced strategies for training and especially pre-training could prove a very valuable direction of further research.

\section*{\uppercase{Acknowledgements}}

This work was partially funded by the Federal Ministry of Education and Research Germany under the project DECODE (01IW21001).

\bibliographystyle{apalike}
{\small
\bibliography{Paper}}

\end{document}